\documentclass[sigconf]{acmart}

\usepackage[normalem]{ulem}

\AtBeginDocument{%
  }

\newif\ifshowrevisions
\newif\ifshowdeletions
\showrevisionstrue
\showdeletionsfalse

\DeclareRobustCommand{\rev}[1]{%
  \ifshowrevisions
    {\color{black}#1}%
  \else
    #1%
  \fi
}

\DeclareRobustCommand{\del}[1]{%
  \ifshowdeletions
    {\color{black}\sout{#1}}%
  \fi
}

\copyrightyear{2026}
\acmYear{2026}
\setcopyright{cc}
\setcctype{by}
\acmConference[UIST '26]{The 39th Annual ACM Symposium on User Interface Software and Technology}{November 02--05, 2026}{Detroit, MI, USA}
\acmBooktitle{The 39th Annual ACM Symposium on User Interface Software and Technology (UIST '26), November 02--05, 2026, Detroit, MI, USA}
\acmDOI{10.1145/3830398.3830707}
\acmISBN{979-8-4007-2856-3/2026/11}

\begin{document}

\title[Can You Say This for Me?]{Can You Say This for Me? Speaking Up by Proxy in Co-Located Discussion}

\author{Yue Shen}
\email{yuesh@vt.edu}
\affiliation{%
  \institution{Virginia Tech}
  \department{Department of Computer Science}
  \city{Blacksburg}
  \state{Virginia}
  \country{USA}
}

\author{Rehema Abulikemu}
\email{rexime@vt.edu}
\affiliation{%
  \institution{Virginia Tech}
  \department{Department of Computer Science}
  \city{Blacksburg}
  \state{Virginia}
  \country{USA}
}

\author{Ryan P. McMahan}
\email{rpm@vt.edu}
\affiliation{%
  \institution{Virginia Tech}
  \department{Department of Computer Science}
  \city{Blacksburg}
  \state{Virginia}
  \country{USA}
}

\author{Yan Chen}
\email{ych@vt.edu}
\affiliation{%
  \institution{Virginia Tech}
  \department{Department of Computer Science}
  \city{Blacksburg}
  \state{Virginia}
  \country{USA}
}

\begin{abstract}
Equal participation in co-located discussion is important for effective collaboration, yet people often hold back when they anticipate negative interpersonal or professional consequences, especially when raising a point requires voicing it themselves. We present SecondVoice, a mixed-reality system that enables people to speak up through an embodied virtual proxy. By separating what is said from who says it, SecondVoice brings hesitant points into the live spoken discussion without putting the speaker on the spot. Using a private overlay, users specify their intent through a structured specification process rather than composing a full utterance. The system reformulates the input and voices it into the conversation through the proxy. \rev{We characterize a design space of participation channels under social risk.} In a \rev{preliminary} within-subject study ($N=16$)\del{comparing SecondVoice with an anonymous text-board baseline }, \rev{we compare the complete SecondVoice system with an anonymous text-board channel} across two group discussion tasks\del{, half }. Half of \del{SecondVoice users activated the channel }\rev{participants reported using SecondVoice }for a point they did not say aloud, compared with 18.8\% for the text board. Proxy-delivered points entered the spoken floor and \del{prompted }\rev{were followed by }multi-turn group engagement, \del{ that text-board posts did not}\rev{which we did not observe after text-board posts}. Participants described the channel as situationally valuable but identified tradeoffs around timing, ownership, and trust in reformulation.
\end{abstract}



\begin{teaserfigure}
  \centering
  \includegraphics[width=\textwidth]{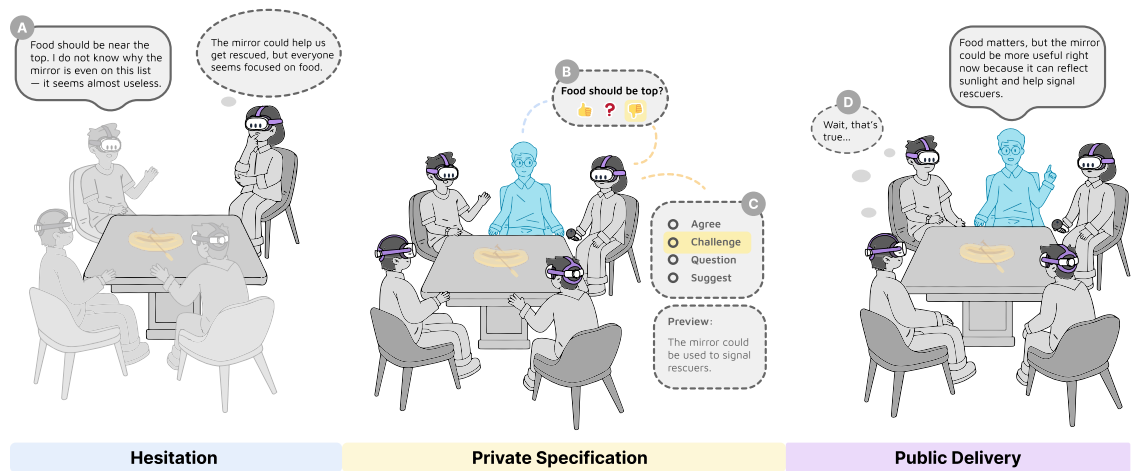}
  \caption{A group ranks survival items after a shipwreck. Everyone agrees food belongs at the top; one participant thinks the mirror is more important but does not want to be the one to say so~(A). The system privately surfaces the active topic with stance options~(B). She indicates disagreement, selects a communicative move, and confirms a \del{phrasing}point through the structured specification interface~(C). A shared embodied proxy raises her point on the spoken floor \del{without revealing who authored it}without naming the author~(D).}
  \Description{Four-panel illustration of the SecondVoice system, depicting a Lost at Sea survival-ranking task. Left panel, labeled Hesitation: four people wearing MR headsets sit around a table; one participant speaks aloud that food should be near the top and the mirror seems useless, while another participant has a thought bubble considering that the mirror could help get rescued but everyone is focused on food. Center panel, labeled Private Specification: SecondVoice surfaces a pulse asking ``Food should be top?'' with three stance buttons; below, the participant sees the specification interface with four communicative move options (Agree, Challenge, Question, Suggest) with Challenge selected, and a preview reading ``The mirror could be used to signal rescuers.'' The proxy figure, rendered in blue, sits at the table. Right panel, labeled Public Delivery: the proxy figure raises its hand and speaks the reformulated point about the mirror reflecting sunlight to help signal rescuers; another participant responds ``Wait, that's true\ldots''\del{ indicating the group has taken up the point}.}
  \label{fig:teaser}
\end{teaserfigure}

\maketitle

\section{Introduction}
Co-located group discussion remains one of the most common settings in which people critique ideas, negotiate tradeoffs, and make decisions together. Research on collective intelligence suggests that more equal participation in discussion is associated with stronger group performance~\cite{woolley2010evidence}. Yet achieving this equality in face-to-face discussion has proved difficult: in classrooms, many students participate minimally in whole-class discussion~\cite{sedova2020silent}; in the workplace, women and lower-status members often speak less despite having relevant expertise~\cite{karpowitz2012gender,nembhard2006making}. Often the problem is not a lack of ideas but a reluctance to voice them. People frequently withhold concerns, disagreement, or suggestions as they fear being judged as uninformed, disruptive, or difficult to work with~\cite{morrison2000organizational,milliken2003exploratory,edmondson1999psychological}. This pressure is especially salient in face-to-face settings, where there are few ways to raise a substantive point without also claiming the floor yourself.

Prior work has largely explored two paths to broaden participation in co-located discussion. One focuses on anonymous or parallel channels, such as backchannels and anonymous text boards, that allow people to participate without speaking up~\cite{mccarthy2005digital,harry2009backchan,bergstrom2009vote,nelimarkka2014field}. While they can broaden participation opportunities, remarks often remain in a side layer that others may overlook or never surface~\cite{harry2009backchan,bergstrom2009vote}. Co-located use can further weaken these protections, since even the act of typing can be visible to others. A second path emphasizes facilitation, nudging people to speak up within the ongoing conversation~\cite{dimicco2007impact,kim2008meeting,samrose2021meetingcoach,hancock2020ai,kadoma2024role}. 
Neither path keeps a point in the live spoken discussion while lowering the social cost of voicing it.

Recent work on agents in group discussion points to a third possibility. Agents now enter live group interaction as facilitators, advocates, or meeting coaches~\cite{houtti2025observe,lee2025amplifying}. Especially when embodied, such agents can take turns in live discussion as co-present speakers in the room. Prior work also suggests that embodiment can strengthen rapport and trust, and improve synergy in group interaction~\cite{shamekhi2018face,kim2024engaged,ma2025nods}. Yet existing systems typically position these agents as independent social actors, which could encourage over-reliance and make some participants feel less needed in the discussion~\cite{johnson2025exploring,johnson2025augmenting}. This leaves an underexplored middle position: a bounded embodied proxy that takes the floor on behalf of hesitant participants, voicing only what they have privately specified. 

We present \textit{SecondVoice}, a mixed reality (MR) system that allows people who hesitate to speak up in co-located discussion to have what they want to say voiced by a shared embodied proxy. MR makes this tractable by combining a private asymmetric interface with a shared embodied speaker in the same physical space. 
Rather than composing a full utterance, users specify their remark at the level of intent through a structured specification process: selecting a communicative move, anchoring it to an active discussion topic, and refining its expression. 
The system grounds that intent in the ongoing discussion, reformulates it into context-appropriate speech, and delivers it through the proxy. This architecture partially separates deciding what to say from publicly bearing the exposure of voicing it.

We compare SecondVoice with a common side channel, an anonymous text board where participants can post to a shared display visible to the group. In a within-subject study across two group discussion scenarios, we examine \del{whether proxy-mediated participation helps hesitant points enter the discussion}\rev{how hesitant points enter the discussion through each channel}, how groups take up those points, and how participants use \del{the }each channel in practice. \del{Our findings suggest that participants more often used SecondVoice for points they were reluctant to voice directly, and that proxy-delivered points were harder to overlook than text-board posts. The proxy channel also reached a broader set of contributors, including quieter participants in the group. Participants described clear tradeoffs around timing, trust, and ownership. }\rev{Participants reported using SecondVoice for points they were reluctant to voice directly. Three observed proxy deliveries were followed by multi-turn, multi-speaker exchanges, which we did not find after text-board posts.}  \rev{Participants described the proxy as reducing the pressure of speaking in their own voice, but also raised concerns about timing, trust, and ownership. They used the proxy for different purposes, and in one case it became a stepping stone back into direct speech.}

This paper contributes: (1)~the design and implementation of a proxy-mediated participation system for socially risky co-located discussion\rev{, together with a design space characterizing tradeoffs among participation channels}, (2)~a structured specification process that lets users specify a remark at the level of intent before the system reformulates and delivers it through a shared embodied proxy, and (3)~\rev{preliminary} empirical findings from a comparative study examining how proxy-mediated participation is used and experienced relative to an anonymous text-based channel.

\section{Related Work}

\subsection{Participation Barriers and Parallel Channels}
A longstanding response to the social costs of speaking up in face-to-face discussion has been to create parallel channels that let participants contribute without immediately taking the spoken floor. In HCI and CSCW, these channels include digital backchannels, anonymous text boards, and low-effort  backchannel signals that support question asking, commentary, and low-interruption participation alongside live discussion \cite{mccarthy2005digital,harry2009backchan,bergstrom2009vote,nelimarkka2014field}. They broaden participation opportunities by creating a lower-pressure route for asking questions, adding commentary, or signaling disagreement.

Most of these systems, however, keep contributions in a separate textual or signaling layer rather than returning them to the live spoken exchange. In \textit{backchan.nl}, for example, audience questions and votes become a ranked public feed, but uptake still depends on a moderator or speaker noticing and voicing them \cite{harry2009backchan}. Low-effort  signal systems such as \textit{Vote and Be Heard} similarly make participation more legible, yet require interpretation before nuanced ideas affect the front-channel discussion \cite{bergstrom2009vote}. Co-located use also complicates the protection these channels offer. McCarthy and boyd note that identifiable handles and traceable logs can weaken anonymity even when contributions are mediated through a backchannel \cite{mccarthy2005digital}. As a result, parallel channels can make contributions easier to overlook and leave their uptake dependent on moderators, instructors, or dominant speakers. They lower direct attribution pressure, but reintegration into the shared spoken floor remains largely unaddressed.

\subsection{Discussion Support and AI-Mediated Expression}
Other systems support participation within the ongoing conversation itself, typically by making speaking-time imbalance visible or by coaching participants toward more balanced interaction. Real-time participation displays such as \textit{Second Messenger} \cite{dimicco2007impact}, wearable sociometric sensing in \textit{Meeting Mediator} \cite{kim2008meeting}, shared turn-taking visualizations such as \textit{Conversation Balance} \cite{li2022conversation}, and post-meeting reflective dashboards such as \textit{MeetingCoach} \cite{samrose2021meetingcoach} address different stages of participation support, but they share a common assumption. Participants still need to adjust their own behavior and publicly voice the contribution themselves. These systems can make imbalance more visible and sometimes support more balanced participation, yet they generally do not change who must publicly voice a socially risky contribution.

AI-mediated communication takes a different approach. Rather than regulating who speaks, it reshapes how a message sounds by augmenting, rewriting, or generating text on behalf of a communicator \cite{hancock2020ai}. AI can alter message form, but doing so also changes how much control users feel they retain. Workplace co-writing studies find that even stylistic assistance can affect perceived agency and ownership \cite{kadoma2024role}. Across both paths reviewed so far, existing systems either regulate participation or reshape message form, but they rarely offer a channel that reintroduces user-authored, socially risky points into the shared spoken floor.

\subsection{Conversational Agents and Embodied Participation}
Agents that participate directly in group discussion have taken on a range of roles. At one end, facilitator agents monitor participation and prompt underrepresented speakers: \textit{Observe, Ask, Intervene} does so in online meetings~\cite{houtti2025observe}, while \textit{ClassMeta} deploys virtual classmates in a VR lecture hall that break silences and offer partial answers to create openings for students~\cite{liu2024classmeta}. Advocate agents go further by generating content: \textit{Amplifying Minority Voices} uses an AI devil's advocate to surface counterarguments in group decision-making~\cite{lee2025amplifying}. At the other end, surrogate agents stand in for the user entirely. \textit{Dittos} creates personalized embodied agents that attend meetings on a user's behalf, matching their appearance and voice~\cite{leong2024dittos}. \rev{Nam et al.\ reconstruct absent attendees so that asynchronous viewers can put questions to them~\cite{nam2025effects}.} Cheng et al.\ explore shared autonomy in voice calls, where an agent handles portions of a conversation while the user multitasks~\cite{cheng2025conversational}. Across this spectrum, existing systems share a common property: the agent either decides what to say on its own or replaces the user altogether. Research on intervention strategy reinforces a related concern. Private, individual prompts from an agent are more effective and better received than public ones~\cite{do2022should}, and high social prominence in an autonomous agent can suppress participants' critical thinking and reduce their sense of being needed~\cite{johnson2025exploring,johnson2025augmenting}. \del{What remains unexplored is an agent role in which the user privately authors a specific point and the agent voices it on their behalf within an ongoing discussion. }\rev{This proxy role remains underexplored in synchronous, co-located group discussion, where one shared proxy can voice points privately specified by present participants.}

Embodiment shapes how such agent contributions land in a group. Embodied facilitation agents receive higher rapport and trust ratings and lead to more balanced turn-taking than disembodied counterparts~\cite{shamekhi2018face}. Virtual agent appearance and behavior further influence engagement, inclusivity, and satisfaction in group settings~\cite{kim2024engaged,ma2025nods}. In social VR, multimodal attention cues help group members notice a new speaker attempting to join the conversation, underscoring how embodied presence supports turn-taking entry in shared virtual spaces~\cite{lee2024may}. These findings motivate using an embodied, spoken proxy to present a user-authored point as a turn in the live spoken discussion. In the autonomous examples above, the embodied agent acts on its own judgment rather than serving as a conduit for a particular user's remark. \rev{Personal avatars can instead mediate a present user's own expression. Do et al.\ render typed text through personalized affective avatars in one-to-one virtual meetings for autistic adults and adults with social anxiety~\cite{do2025exploring}. Each avatar remains associated with the user whose text it voices.}

SecondVoice\del{addresses this gap. Rather than acting autonomously or replacing the user, it positions the agent as a bounded proxy. } uses one shared, bounded proxy \rev{in co-located discussion. Unlike summarizers that aggregate views, each delivery in our study voiced one participant-confirmed point. MR gives each participant a private input layer while rendering the same co-present proxy to the group, a combination that a shared display alone would not provide.} The next section formalizes the design space that motivates this choice.

\section{Participation Channels Under Social Risk}
Different ways of participating in co-located discussion impose different trade-offs. Drawing on Media Richness Theory~\cite{daft1986organizational} and prior work on backchannels and parallel participation~\cite{mccarthy2005digital,harry2009backchan,bergstrom2009vote,nelimarkka2014field}, we compare existing participation channels along three properties that matter when social risk is present: how much direct exposure they impose on the person raising a point, how precisely they let that point be specified, and whether it enters the shared spoken floor at all. \rev{The spoken floor is the group's shared conversation, and a point enters it as a turn others can answer.} Direct speech supports the highest expressive specificity among real-time channels and places a point directly into the spoken floor, but it ties that point to the speaker's public self-voicing in the moment. Anonymous text boards and backchannels can lower direct exposure while still supporting nuanced expression, yet they typically keep remarks in a side layer that the group may overlook. Low-effort  reactions and polls reduce the burden further, but at the cost of saying much less. \del{Facilitation and nudging can encourage people to speak, but they function as supportive layers around these channels rather than as channels for expressing a specific point themselves. }\rev{Nudging can encourage people to speak, but it does not carry a specific point itself. A human read-aloud relay can bring a submitted point into the spoken discussion, but it still requires a low-exposure submission route and a facilitator to relay it.}

These trade-offs reveal an underserved region in co-located discussion (Table~\ref{tab:channel_comparison}). Existing channels tend to offer either lower exposure outside the spoken floor, or entry into the spoken floor at the cost of direct exposure. What remains underserved is a channel that stays expressive enough for a nuanced point, lowers the direct exposure of raising it, and still lets it enter the live spoken discussion. Filling this gap is not only a matter of where a point appears. It also depends on \textit{who voices it}: if the point could be spoken by someone other than the author, the coupling between content and exposure can be partially broken~\cite{goffman2017interaction}.

\begin{table}
  \centering
  \caption{Participation channels under social risk. Existing channels tend to offer either lower exposure outside the spoken floor or spoken-floor entry at the cost of direct self-voicing. The underserved region is a channel that remains expressive, lowers direct exposure, and still enters the spoken floor.}
  \Description{A low-fidelity comparison table positioning participation channels by exposure, specificity, and whether they enter the spoken floor.}
  \label{tab:channel_comparison}
  \small
  \setlength{\tabcolsep}{2pt}
  \begin{tabular}{p{1.2in}ccc}
    \hline
    Channel & Direct exposure & Specificity & Spoken floor \\
    \hline
    Direct speech & High & High & Yes \\
    Anonymous text board / backchannel & Lower & Medium--High & No \\
    Reactions / polls & Lower & Low & No \\
    Facilitation / nudging & Medium & Low & Indirectly \\
    Autonomous agent / surrogate & Lower for user & Variable & Yes \\
    SecondVoice / bounded proxy & Lower for author & Medium--High & Yes \\
    \hline
  \end{tabular}
\end{table}

The table's last two rows point to one possible answer. As reviewed in Section~2.3, agents can now take the spoken floor in live discussion, but in roles that either determine the content themselves or replace the user entirely. What remains underexplored is the narrower role shown in the final row: a shared, bounded proxy that voices only what a user has specified and confirmed.

Delegating delivery to a proxy partially separates content from exposure, but the separation itself introduces a tradeoff between timeliness and authorship. A system that generates and delivers a point autonomously can keep up with the conversation, but the user loses authorship over what is said. A user who composes the full utterance for a relay retains authorship, but the conversation may move on before the point is ready. \rev{Here, an utterance means the complete wording to be spoken rather than the underlying point.} Between these extremes lies a region where the user specifies intent through structured selection rather than open-ended composition, \del{retaining authorship while reducing the time needed to produce a ready point}retaining authorship over the point while reducing how much the user must compose from scratch. \textit{SecondVoice} explores whether this middle ground can preserve enough user authorship while staying timely enough for live discussion.

\section{SecondVoice}
\begin{figure*}[t]
    \centering
    \includegraphics[width=1\linewidth]{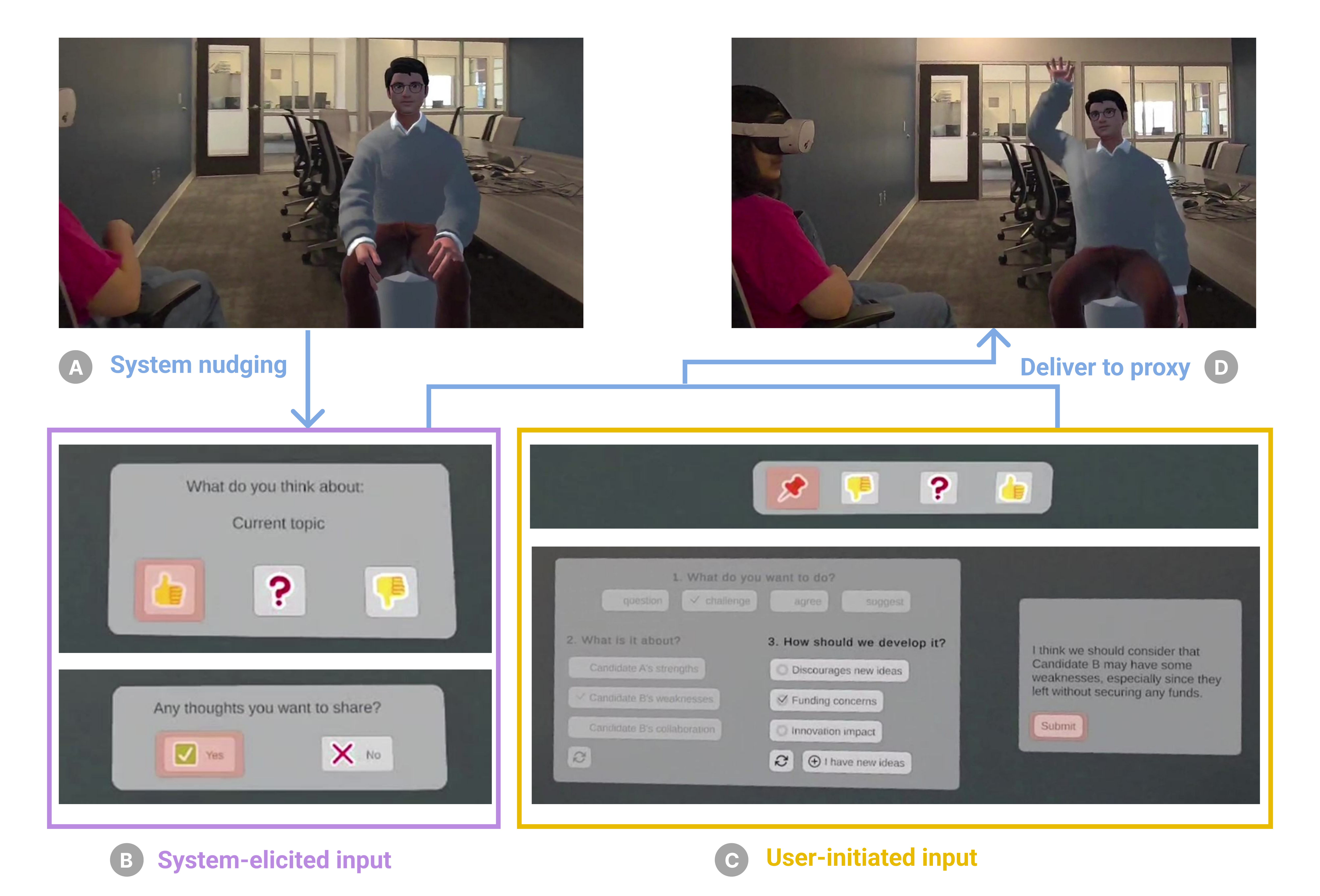}
    \caption{The SecondVoice interaction flow. (A)~The system monitors the discussion and may initiate contact through a pulse or nudge. (B)~System-elicited input: a pulse surfaces the current discussion topic with stance options (agree, question, disagree); a nudge asks whether the participant wants to share a thought. (C)~User-initiated input: the participant opens the structured specification interface, selects a communicative move, anchors it to a discussion topic, and browses candidate points. After the participant confirms the point, the system reformulates it against the current discussion. (D)~The proxy raises its hand and delivers the specified point on the spoken floor.}
    \Description{Four-part figure showing the SecondVoice system. Top-left (A): the proxy sits at a table during discussion. Bottom-left (B, purple border): two system-elicited panels---a pulse showing the current topic with three stance buttons and a nudge asking whether the participant wants to share, with Yes and No buttons. Bottom-right (C, yellow border): the full structured specification interface with four attitude icons, three-step selection (communicative move, discussion topic, candidate points), and a text preview with Submit button. Top-right (D): the proxy raises its hand to deliver the point. Arrows show the flow from A to B to C to D.}
    \label{fig:system_figures}
\end{figure*}

This section describes how SecondVoice realizes the proxy-mediated participation channel introduced above. The system organizes participation into three stages: specification, \del{grounding, and delivery }\rev{contextual reformulation}, and \rev{proxy delivery} (Figure~\ref{fig:system_figures}). \rev{Discussion grounding runs continuously across these stages, providing context for candidate generation, reformulation, and delivery timing.} Together, these stages address the gap identified in Section~3: they let a user specify an expressive point, lower the direct exposure of raising it, and still place it on the shared spoken floor. We first illustrate the interaction through a walkthrough, then describe each stage and the current implementation.

\subsection{Walkthrough}
Imagine a small-group discussion about selecting a university president. Three colleagues are converging on Candidate~A, while you think Candidate~B has a strength the group has overlooked. Raising that concern directly would place you visibly at odds with the emerging majority. As the discussion continues, you open a private overlay on your headset and place a pin on the moment where the group dismissed Candidate~B's experience. The pin carries an attitude tag, \textit{disagree}, recording your stance toward the moment. Through the private overlay, visible only to you, you then open the structured specification process. You select a communicative move, \textit{challenge}, anchored to the current candidate-selection topic. The system presents three candidate points; you scroll through them using the controller thumbstick. One option captures your concern. You press the trigger to confirm the selected point. The system then reformulates it against the current discussion before delivery.

The system prepares the point for spoken delivery through Bob, the proxy, a shared embodied character visible to all participants in the room. Bob is rendered as a co-present speaker sitting at the table. When a brief pause arises, Bob raises his hand to signal, then speaks the point directly into the conversation in his own voice, without announcing that someone has something to add or \del{disclosing who authored the remark}naming the author. The group hears him as another participant entering the discussion. When a colleague asks the proxy to elaborate, it offers only a brief clarification rather than developing the point on its own. To continue, you reopen the private interface and select or refine a follow-up for Bob to deliver.

\subsection{System Overview}
SecondVoice combines two coupled layers: a private, per-user interaction layer for specifying points and a shared embodied proxy that can speak them into the room (Figure~\ref{fig:system_figures}). The private layer runs on each participant's MR headset; a centralized backend maintains the shared discussion state that connects them.

\rev{Discussion grounding runs throughout the three stages.} A rule-based bookkeeping layer maintains a running account of recent transcript turns, question cues, active pins, and coarse topical anchors from the transcript, accumulating and organizing this data as it arrives without requiring language-model calls.

Periodically, the system runs an observation pass that extracts the current discussion topic, open issues, topical anchors, and speaker-attributed stance signals from the recent transcript. A profile update step turns those signals into per-user stance summaries and active concerns. These extracted elements directly feed the specification interface: the topic options a user sees when anchoring a point, and the candidate points generated for a given move-topic pair, both draw on this evolving discussion state. Users therefore see options tied to what the group is currently discussing rather than generic or stale prompts.

The discussion state also supports post-confirmation reformulation, while per-user participation signals help the system judge whether a specified point remains timely enough for later delivery. The following subsections describe each stage in detail.

\subsection{Participation Entry and Specification}
The first stage addresses the specificity dimension: it lets a user specify a nuanced point without having to voice it in real time. It opens the channel and turns a brief reaction into a structured point specification. A user may place a \textit{pin} on a moment in the discussion that feels worth returning to, such as a claim they disagree with, a question they want to raise, or a concern they are not yet ready to voice aloud. Each pin carries an optional attitude tag (agree, question, or disagree) that captures the user's stance toward the moment, or the user can simply pin without indicating a stance. This lets the user begin by recording a reaction before fully articulating it. The attitude tag also pre-fills the default communicative move when the user later opens the specification interface, reducing the number of decisions needed to begin.

The system also monitors each user's participation state, including cumulative and recent speaking activity, time since last speaking, and unresolved pin history. Two private prompting mechanisms use these signals to offer optional entry points without generating content or speaking on the user's behalf. A \textit{pulse} surfaces a concrete discussion anchor (for example, the current topic and its active stance options) when a user has been silent while the discussion remains substantive. It shows the user what the group is talking about and offers a one-tap entry point into the specification process. A \textit{nudge} is a private text reminder (for example, ``You pinned some ideas earlier, want to open the pad?''), triggered when a user has unresolved pins or has remained silent for an extended period. The system checks for pin reminders first, then pulses, then generic nudges, so that the most contextually specific prompt takes priority. Both are private and dismissible.

The structured specification process is the primary mechanism for this stage. Rather than asking the user to type a finished sentence, it decomposes what would normally be a single, cognitively heavy act of real-time speech formulation into a sequence of guided selections. Speech production models describe real-time formulation as cognitively demanding, requiring simultaneous planning, word retrieval, and encoding under temporal pressure~\cite{levelt1993speaking}. The structured specification process replaces open-ended generation with recognition and selection.

\begin{figure}[t]
  \centering
  \includegraphics[width=\columnwidth]{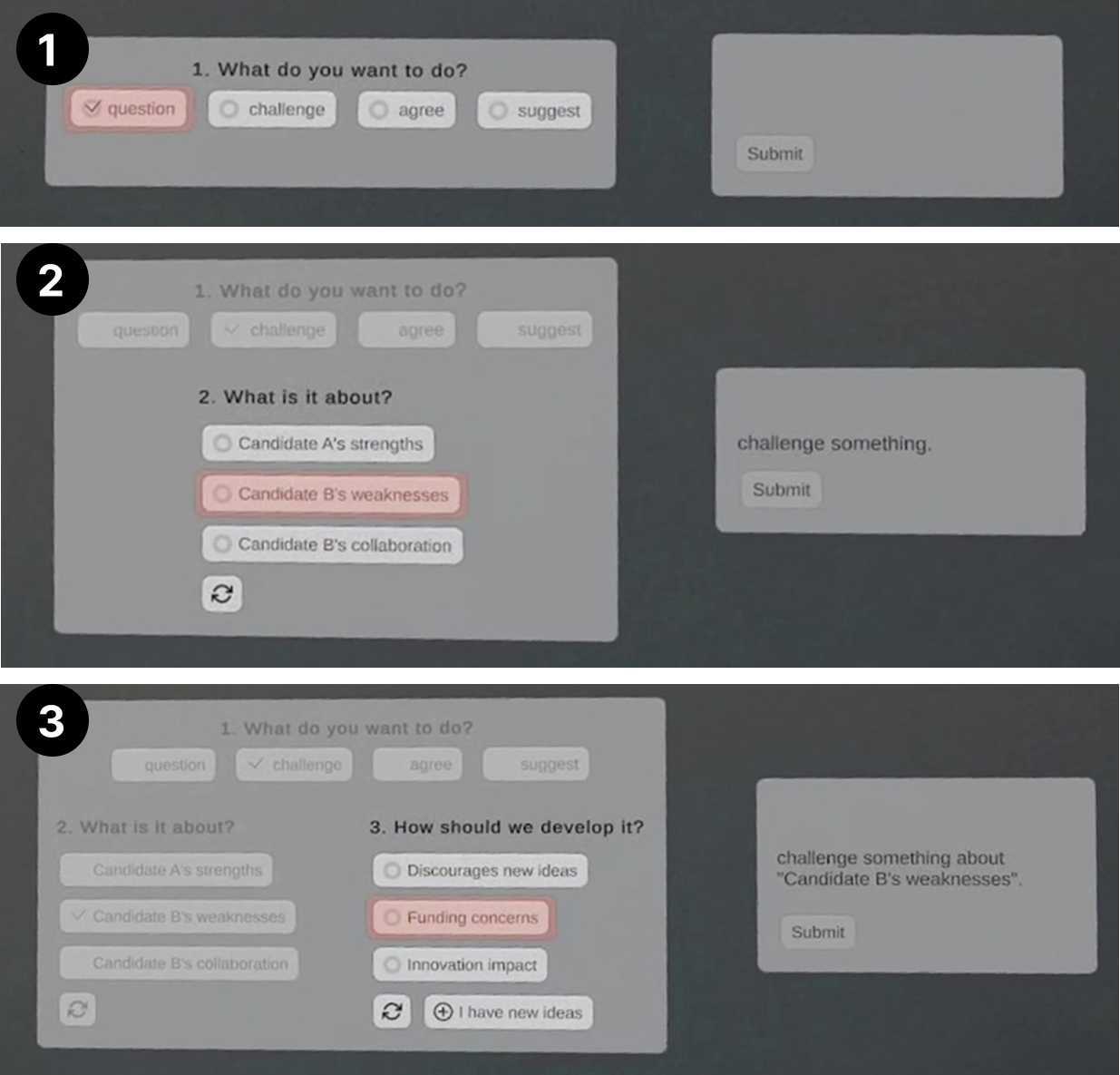}
  \caption{The three-step structured specification interface. (1)~Select a communicative move: question, challenge, agree, or suggest. (2)~Anchor the move to an active discussion topic. \del{A typed fallback (right) is available at each step. }(3)~Browse and refine candidate points generated for the selected move--topic pair; a refresh button produces alternatives.}
  \Description{Three vertically stacked panels showing the specification interface. Panel 1 shows four move buttons (question selected). Panel 2 adds three discussion-topic options with one highlighted and a text preview panel. Panel 3 adds candidate point options with a refresh button, and the preview panel shows the accumulated specification.}
  \label{fig:csi}
\end{figure}

The process proceeds in three steps (Figure~\ref{fig:csi}). It begins with selecting a \textit{communicative move}: questioning, agreeing, challenging, or suggesting. These four moves cover the primary speech acts relevant to group discussion: support and opposition map to assertive and expressive acts, while questioning and suggesting map to directive acts~\cite{searle1969speech, core1997coding}. The move is then anchored to an \textit{active discussion topic} drawn from the system's evolving discussion state, situating the point in the current conversation rather than in the abstract. From there, the user browses \textit{candidate points} that the system generates for the selected move and topic. Three candidates are shown at a time to avoid choice overload~\cite{hick1952rate}. If none of the candidates match the user's intent, they can refresh: without a selection, the refresh produces a new random set; with a candidate selected, the refresh anchors on the selected item and generates similar alternatives, allowing the user to iteratively converge on the point they want to make. The user confirms the selected candidate as the point to convey.

The entire specification process uses the Quest controller's thumbstick and trigger. The user moves a highlighter across options with the thumbstick and confirms a selection with the trigger. Compared to ray-casting or direct hand tracking, thumbstick input requires minimal hand movement: users can rest their hand naturally on an armrest or their lap while operating the interface\del{, making the specification process difficult for other participants to observe}. \rev{This keeps hand movement small and makes the interaction less conspicuous}.

\subsection{\del{Discussion Grounding and Intent Interpretation}\rev{Contextual Reformulation}}
\del{The second stage interprets the specified point against the current discussion.}

\ifshowrevisions
\textbf{\del{Reformulation and Proxy Delivery}}\par
\fi
The \del{third stage addresses the remaining two dimensions by reformulating, timing, and delivering the specified point through the proxy, placing it on the shared spoken floor while lowering the author's direct exposure}\rev{second stage reformulates the confirmed point against the current discussion}. Once a point has been grounded in context, the system reformulates it into a spoken utterance suited to the current moment, adapting tone and phrasing to the ongoing discussion. The proxy \del{preserves }\rev{is instructed to preserve} the user's intended point while adapting phrasing to the live conversation, without introducing a new stance or elaborating beyond what the user specified. \rev{Because the system reformulates the point after confirmation, the delivered wording can differ from the selected candidate.}

\subsection{\rev{Proxy Delivery}}

In the third stage, the proxy delivers the reformulated point in its own voice, as though it were offering the remark itself. It does not announce that someone has something to add, \del{disclose hidden authorship}\rev{name the author}, or frame the point as relayed from another participant. To the rest of the group, the proxy sounds like a co-present participant entering the discussion. Because the point enters the conversation through the proxy rather than as a flagged message from a specific participant, \del{the author's exposure remains low}\rev{the author does not have to voice it directly}. All users share a single proxy\del{, so the group cannot determine who authored a given delivery from the proxy's behavior alone}. \rev{The design aims for lower exposure rather than anonymity. Behavioral or contextual cues, such as who was not speaking when the proxy delivered a point, could still suggest who contributed it.}

Delivery timing is treated as a design problem. The system places pending speech into a queue and checks whether the point remains timely, whether the topic has shifted, and whether a similar point has already entered the room. The proxy then waits for a brief conversational pause before speaking. Waiting for a pause can separate interface operation from delivery, but the point may become less timely as the discussion moves on. Prior research identified a ``standard maximum silence'' of approximately one second in conversation, beyond which participants treat the floor as unoccupied~\cite{jefferson1989preliminary}. The current implementation uses a 1.2\,s silence threshold, set just above this boundary and confirmed through pilot testing.

When the proxy is directly addressed by another participant, its conversational competence remains bounded. It may provide a brief clarification or bridge response tied to the delivered point, but \del{substantive follow-up is redirected back through the private channel. After each }\rev{it cannot answer substantively on the author's behalf}. \rev{If the author reopens the private interface after a recent delivery, a \textit{follow-up panel} offers candidate clarifications based on the original point and subsequent discussion. The author can select or refresh these options or return to the full specification process. A selected option is delivered directly as a new proxy turn.} After each primary delivery, the system privately asks the author whether the delivery was accurate, partially correct, or not their view. The system stores the response and uses it to inform later reformulations for that participant.

\begin{figure}[t]
  \centering
  \includegraphics[width=\columnwidth]{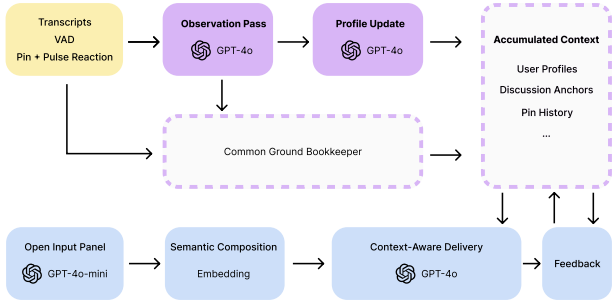}
  \caption{Backend pipeline architecture. Top: a periodic monitoring loop uses GPT-4o to extract discussion state and per-user stance summaries; a rule-based bookkeeping layer tracks question cues and pin rankings without LLM calls. Bottom: when a user opens the specification interface, GPT-4o-mini generates \del{three-step options }the initial three-step specification screen (communicative move $\times$ discussion topic $\times$ candidate points). A sentence-embedding model supports anchored refresh and contribution deduplication. Context-aware delivery reformulates the specified point into proxy speech via GPT-4o without naming its author; post-delivery feedback can inform later reformulations.}
  \Description{Two-part pipeline diagram. The top half shows a periodic monitoring loop: speech-to-text feeds into an observation pass and profile update, with a rule-based bookkeeper accumulating context. The bottom half shows the user-triggered contribution flow: structured option generation, anchored refresh and similarity checks, and context-aware proxy delivery with a post-delivery feedback loop.}
  \label{fig:pipeline}
\end{figure}

\subsection{Implementation}
The current prototype uses a split client-server architecture (Figure~\ref{fig:pipeline}). Each participant wears a Meta Quest~3 headset running a Unity client that handles private overlays, microphone capture, proxy rendering, and shared virtual objects. Each headset connects to an async Python backend (FastAPI) via two per-user WebSocket channels: a binary audio stream and a JSON control channel. The backend concurrently manages per-user speech-to-text streams (Deepgram Nova-3, one dedicated session per user for attributed transcription), a periodic observation pass that extracts discussion state from the live transcript, specification-interface sessions for each user, a delivery queue with timeliness and deduplication checks, and a participation monitor that drives pulse and nudge decisions. It also maintains per-user profiles containing stance summaries and active concerns. Language-model calls use a triple-model split. GPT-4o handles observation passes, subsequent candidate generation and refresh, semantic adaptation, and follow-up bridging. GPT-4o-mini handles initial three-step specification-screen generation (reducing first-screen latency) and the substantiveness check used to gate pulses. A sentence embedding model (all-MiniLM-L6-v2) supports candidate similarity scoring during anchored refresh. Proxy speech is synthesized via Deepgram Aura text-to-speech (aura-2-aries-en). The proxy is a shared Ready Player Me character rendered at a fixed calibrated position on each headset, so all participants perceive it at the same physical location while retaining personalized private overlays. \rev{The backend also included a path for clustering similar pending contributions. Each delivery in the study was based on one participant's contribution, so we did not evaluate aggregation across participants.}

\section{User Study}
We conducted a within-subject study to \del{evaluate whether SecondVoice supports participation }\rev{examine SecondVoice as a participation channel} under social risk in co-located discussion. \del{We asked three questions: whether the proxy channel helps hesitant points reach the spoken floor more readily than an anonymous text-board baseline, whether those points are taken up by the group, and how participants use each channel in practice. Both conditions use }\rev{We chose an anonymous text-board as the baseline because it could carry a specific point without requiring its author to voice it directly, while keeping that point in a visual side channel rather than placing it on the spoken floor. We asked how the two channels surfaced hesitant points, how groups took up those points, and how participants used each in practice.} Both conditions used MR headsets and private visual interaction layers.

\subsection{Participants}
Sixteen participants (10 male, 6 female) took part in the study, organized into four groups of four. All were undergraduate or graduate students recruited via institutional email, and study eligibility required them to be at least 18 years old. A screening questionnaire asked about self-reported discussion participation tendency and prior VR or MR experience. Each group was composed to balance two participants who described themselves as more vocal and two who described themselves as quieter in group discussion. \del{Most participants had used VR only once or twice before the study. }\rev{We matched screening responses for 16 participants; their prior VR or MR experience ranged from no prior use to frequent use (Table~\ref{tab:participant_profile}).} Each participant received \$25 USD for a session lasting approximately 90 minutes.

\begin{table}[t]
  \centering
  \small
  \setlength{\tabcolsep}{3.5pt}
  \caption{\rev{Participant profile. Speak up and Hold back summarize two screening items about usually speaking up and sometimes holding back relevant ideas (1 = strongly disagree, 5 = strongly agree). VR/MR abbreviations are Occ. for occasionally, Freq. for frequently, and 1--2 for once or twice.} \rev{Unsaid-use SV and ATB give each participant's questionnaire answer to whether they used that condition's channel for something they did not say aloud. Two participants completed the screening questionnaire twice; the later response is reported.}}
  \Description{Table listing the 16 participants (P1 to P16) with, for each, self-reported gender, prior VR or MR experience, two five-point screening ratings for usually speaking up and sometimes holding back relevant ideas, and yes or no answers for whether they used the SecondVoice channel and the anonymous text board for something they did not say aloud.}
  \label{tab:participant_profile}
  \begin{tabular}{llccccc}
    \toprule
     & & Prior & Speak & Hold & \multicolumn{2}{c}{Unsaid-use} \\
    \cmidrule(lr){6-7}
    ID & Gender & VR/MR & up & back & SV & ATB \\
    \midrule
    P1 & F & Occ. & 4 & 4 & No & No \\
    P2 & F & 1--2 & 2 & 4 & Yes & No \\
    P3 & M & Occ. & 4 & 2 & No & No \\
    P4 & F & 1--2 & 3 & 4 & Yes & Yes \\
    P5 & F & Occ. & 3 & 4 & No & No \\
    P6 & M & Freq. & 4 & 2 & Yes & No \\
    P7 & F & Freq. & 2 & 3 & Yes & No \\
    P8 & M & 1--2 & 4 & 2 & Yes & Yes \\
    P9 & M & 1--2 & 3 & 2 & No & No \\
    P10 & M & 1--2 & 5 & 3 & No & No \\
    P11 & M & 1--2 & 3 & 2 & No & No \\
    P12 & M & 1--2 & 3 & 2 & Yes & Yes \\
    P13 & M & Occ. & 4 & 1 & No & No \\
    P14 & M & Occ. & 4 & 4 & No & No \\
    P15 & M & Never & 5 & 2 & Yes & No \\
    P16 & F & 1--2 & 4 & 4 & Yes & No \\
    \bottomrule
  \end{tabular}
\end{table}

\subsection{Methods}
Each group experienced both conditions across both tasks, with condition order and task assignment fully counterbalanced across the four groups. In the \textit{SecondVoice condition}, participants \del{had access to the full participation channel }\rev{used the structured specification and proxy-delivery channel} described in Section~4. In the \textit{anonymous text-board condition}, participants \del{typed messages into a private input field}\rev{selected keys on a private, in-headset virtual keyboard with a controller}. Submitted text appeared anonymously on a shared board visible to all group members. To reduce the social signal of visibly typing during a live discussion, a virtual tabletop overlay \del{occluded each participant's physical keyboard area; when }\rev{masked participants' views of one another's hands.} When a participant began typing, \del{the overlay in front of them became transparent only in their own view, letting them see their hands while remaining visually unchanged to others}\rev{an opening in their own view gave access to the virtual keyboard, while other participants continued to see the tabletop over the typist's hand area}. This condition also included AI text-completion support \del{to reduce MR typing friction. }\rev{that expanded short entries and reduced how much text participants had to enter in MR}. \rev{The two complete channels differ in specification and input, authoring effort, embodiment, output modality, timing, and spoken-floor entry. The comparison is informative at the channel level, but these bundled differences confound attribution to any single component.} Each discussion round used a task designed to create natural opportunities for socially risky contribution.

Task 1: \textit{Selecting the University President}. This hidden-profile task was adapted from prior group decision-making work~\cite{nicholson2021ve}. Participants evaluated three candidates using a mix of shared information and participant-specific private information, creating information asymmetry and encouraging the introduction of privately held, potentially unpopular, or discussion-shifting points.

Task 2: \textit{Lost at Sea}. This is a common survival ranking exercise often used in inclusive-meeting research~\cite{houtti2025observe}. Groups collaboratively ranked 15 survival items and had to reach a single consensus ordering, creating disagreement about priorities and natural openings for critique, challenge, and minority positions. 

Each session began with a training phase in which participants were introduced to both participation channels. Participants practiced using SecondVoice to express thoughts, and composing messages for the anonymous text board for about 5 minutes each. \rev{They were told that proxy-delivered points came from participant-confirmed contributions, and the contributor would not be named.}

Participants then completed two discussion rounds, one per condition. For each task, participants first completed 5--8 minutes of individual preparation, reviewing the task materials. They then engaged in 10--15 minutes of group discussion. After the first round, participants took a 10-minute break to complete a post-condition survey and prepare for the next task. After both rounds, we conducted one-on-one semi-structured interviews with each participant, lasting approximately 10 minutes, to compare their experiences across conditions, understand their participation strategies, and identify how they used the channels.

\subsection{Measures and Analysis}
Four data sources informed the analysis. Backend logs recorded pins, submissions, proxy deliveries, follow-up deliveries, board posts, and task events, providing the primary record of who used each channel, when, and how often. Audio transcripts anchored these events in the conversation. Uptake was assessed primarily through in-session observation by the research team, who noted whether surfaced points were noticed, responded to, and carried forward in the spoken discussion. Transcripts were used to verify and contextualize these observations\del{, not as the primary source}.

After each condition, participants completed a post-condition questionnaire adapted from the NASA Task Load Index~\cite{hart1988development}, Davison's meeting assessment instrument~\cite{davison1999instrument}, and supplemented with custom items targeting channel-specific experience. \del{Primary paired items }\rev{The 13 primary paired items} measured hesitation, perceived safety, barrier reduction, influence, integration, observability, willingness to reuse the channel, and outcome alignment\del{on 7-point Likert scales}. \rev{Responses used 7-point Likert scales (1 = strongly agree, 7 = strongly disagree)}. Paired binary items asked whether participants had withheld thoughts and whether they had used the channel for something they did not say aloud. \del{For those who reported using a channel, follow-up items probed fidelity, ownership, expressive sufficiency, and effort. }\rev{Follow-up items probed fidelity, ownership, expressive sufficiency, and effort. Only participants who reported using a channel for something they did not say aloud answered them (anonymous text board $n=3$; SecondVoice $n=8$).} \rev{Questionnaire comparisons used participants as the paired unit.} We analyzed the primary Likert items with two-sided Wilcoxon signed-rank tests and Benjamini--Hochberg correction\del{, the binary items with exact McNemar tests, and treated the conditional follow-up items descriptively given their small and uneven counts. } \rev{across the 13-item family}, controlling the false discovery rate (FDR). For the two binary items, we used exact McNemar tests. Conditional follow-up items were treated descriptively because their samples were small and uneven.

Semi-structured interviews followed both discussion rounds. \del{We reviewed the transcripts with attention to }\rev{One researcher conducted a thematic analysis of all 16 interview transcripts.} The analysis examined when participants activated each channel, how they experienced proxy delivery relative to the text board, and what tradeoffs shaped channel use in practice. \rev{After a broad reading of the corpus, the researcher used focused coding to develop a codebook and themes. Participant memos and a claim-evidence map recorded supporting and conflicting cases.} Interview excerpts reported below were lightly edited for readability (filler words removed); square brackets mark added or substituted text.

\section{Results}

Our results combine backend traces, transcript-informed observation, questionnaire responses, and post-study interviews. Across the logged SecondVoice sessions, participants placed 18 pins \rev{and produced 13 proxy deliveries: 10}\del{, confirmed 13 proxy deliveries} through the structured specification process and 3 follow-ups. Across the logged anonymous text-board sessions, participants submitted 6 typed contributions to the shared board.

\subsection{Proxy Use and Participation Patterns}

\begin{figure}[t]
\centering
\includegraphics[width=\columnwidth]{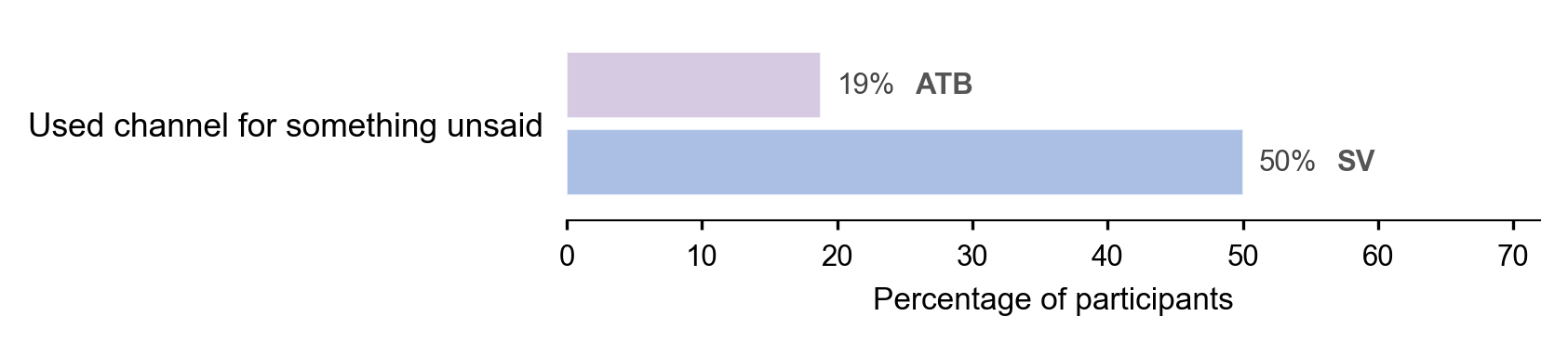}
\Description{Grouped bar chart comparing the two conditions on the share of participants who reported using that channel for something they did not say aloud. The SecondVoice bar is higher than the anonymous text-board bar.}
\caption{Percentage of participants who reported using the channel for something they did not say aloud. Half of \del{SecondVoice users activated the channel }\rev{participants reported using SecondVoice} for an unsaid point, compared with 18.8\% for the anonymous text board (exact McNemar $p=.0625$, $N=16$).}
\label{fig:questionnaire}
\end{figure}

\paragraph{\del{SecondVoice was more often used}\rev{More participants reported using SecondVoice} for points \del{participants}they hesitated to voice directly.}
Half of the participants (8 of 16) reported using SecondVoice to express something they did not say aloud, compared with 18.8\% (3 of 16) for the anonymous text board (Figure~\ref{fig:questionnaire}). \del{All five discordant pairs favored SecondVoice }\rev{The paired difference did not reach statistical significance} (exact McNemar $p=.0625$) \rev{ and is reported as a descriptive trend}. \del{The difference remains trend-level, but the direction is consistent: when participants did use a mediated channel for something they would not have said directly, it was more often SecondVoice. }\rev{P6 explained that proxy speech ``doesn't feel like it's coming from you directly,'' reducing the pressure of raising a point. P7 valued the proxy channel's lower attribution risk and interactivity.}

\paragraph{\del{Quieter participants used the channel to enter the discussion.}\rev{Participants used SecondVoice selectively and for different purposes.}}
The backend logs show \del{that proxy-mediated participation reached participants across the speaking distribution. Across all user-initiated channel uses in}\rev{user-initiated channel use from both halves of the within-session speaking distribution. Of 19 channel-use events across} both conditions (13 proxy deliveries and 6 board posts), 10 \del{of 19 }came from participants in the lower half of their group's speaking distribution for that session, including 5 of the 13 proxy\del{-spoken} deliveries. \del{In the strongest case, }In one survival-ranking session, the participant who spoke least \del{in a survival session produced 4 proxy deliveries through the structured specification process}\rev{(P16) was among the heaviest users of the channel}. \del{In one transcript-backed hidden-profile session, the quietest participant }\rev{The quietest participant in a transcript-backed hidden-profile session (P7) }queued and delivered a question through the proxy despite speaking almost nothing aloud. \del{In both conditions, m}Most participants still preferred to speak directly \del{(94\% ATB, 81\% SecondVoice).}\rev{(15 of 16 in the text-board condition, 13 of 16 with SecondVoice),}\del{T} and the proxy channel was activated selectively\del{, but when it was, it reached participants who otherwise said little on the spoken floor}. Among the proxy deliveries with communicative move labels, \del{quieter participants favored questioning, challenging, and suggesting moves, while more active participants were the only ones who used the proxy for agreement, suggesting the channel served different communicative functions depending on how much a participant was already speaking}\rev{participants used the proxy to question, challenge, suggest, and agree}. \rev{In another hidden-profile session, P12 first said aloud that Candidate C looked like the ideal choice, then used an agreement turn through the proxy to reinforce the same position. Altogether, five participants produced the 13 logged proxy deliveries, four each from P6 and P16, three from P8, and one each from P7 and P12. The six board posts came from P8 (one), P12 (two), and P16 (three).}

\subsection{Group Uptake}

\paragraph{\del{Proxy-delivered points prompted related continuation.}\rev{Some proxy-delivered points were followed by related discussion.}}
Of the 13 proxy deliveries across the three active SecondVoice sessions, 4 \del{prompted } were followed by substantively related human turns within 20 seconds, and 3 of those developed into multi-turn, multi-speaker exchanges. In the strongest episode, the proxy raised a concern about water security in a survival-ranking task. Four different speakers immediately began calculating water quantities and debating collection methods, producing 6 topically related turns. In a hidden-profile session, the proxy asked whether Candidate~C's community-impact record set them apart. The group turned to evaluating whether C was the strongest candidate. In a later exchange from the same task, \del{one participant }\rev{P5 }drew on the proxy's earlier point: ``I think what, like, Bob said that he is pretty good at, like, funding.'' These episodes\del{do not isolate delivery modality from content quality. A well-timed, relevant point may attract follow-on regardless of who voices it. What the transcript traces do show is that proxy-delivered points } \rev{show the complete SecondVoice channel entering} the spoken\del{floor in a way that invited continued engagement, something the board did not produce in comparable form. } \rev{discussion.} \rev{We did not observe a comparable spoken sequence after the six board posts.}

\paragraph{\del{Board posts were }\rev{Some participants found board posts} easier to overlook.}
\del{The anonymous text board produced no comparable spoken-floor event. }\del{Participants repeatedly }\rev{Three participants (P12, P14, and P16)} described the board as something they could acknowledge briefly and move past. As P12 put it: ``The text on the group board is pretty easy to overlook \ldots when the agent speaks \ldots you can't really overlook it.''

When the proxy speaks, it claims a turn and the group reorganizes around it; a board post stays in a parallel layer that others can skim without changing the conversational flow. \rev{P11 offered a different view, expecting visible board text to give the group a point to acknowledge.}

\subsection{Participant Experience and Tradeoffs}

No paired Likert item reached significance after Benjamini--Hochberg correction across 13 primary measures (1\,=\,strongly agree, 7\,=\,strongly disagree; full response distributions in Appendix, Figures~\ref{fig:all_likert1} and~\ref{fig:all_likert2}). We report descriptive trends as context. \del{SecondVoice was rated as providing a more workable way to participate without speaking }\rev{SecondVoice drew slightly more agreement on offering a workable way to participate without speaking} ($M=2.81$ vs.\ ATB $M=3.56$)\del{and as lowering the participation barrier more } \rev{and on lowering the participation barrier} ($M=3.31$ vs.\ $3.75$).\del{ Both conditions were rated similarly } \rev{The conditions were close} on perceived safety, social exposure, and outcome quality. Among the follow-up items answered only by participants who used the channel for something unsaid (SecondVoice $n=8$, ATB $n=3$), \del{SecondVoice users reported higher agreement that the process took mental effort (}\rev{the mean mental-effort rating was} $M=2.50$\del{vs.\ $3.67$), consistent with interview reports of waiting for LLM-generated candidates at each specification step. As secondary task-level context, both } for SecondVoice and $M=3.67$ \rev{for the text board.} Both survival-ranking groups under SecondVoice improved their collective ranking relative to the mean of members' individual rankings ($+14.5$ and $+8.0$ points closer to the expert solution), while neither text-board group did ($-4.5$ and $-6.0$). In the hidden-profile task, three of the four groups chose Candidate~C and one chose Candidate~B. With only two sessions per condition per task, we treat these outcomes as descriptive context for the interview findings that follow.

\paragraph{\del{The }\rev{Participants had mixed experiences with} the structured specification process\del{helped participants formulate remarks.}.}
\del{Several participants described the three-step process as doing more than transmitting a pre-formed thought. }P12 found it faster than composing text from scratch: ``it broke it down really fast \ldots the process was much faster compared to the second one where you actually had to type things out.'' \rev{P8 said the three candidates helped shape an initial thought, and P16 found the options useful for brainstorming in an unfamiliar scenario.} The process did not always match intent: P16 tried refreshing candidates repeatedly but found them too similar: ``it comes up very similar things again and again \ldots it's really hard to put some my ideas directly.'' \del{When generated candidates did not align with a participant's intended point, the process became a barrier rather than a scaffold.}

\paragraph{\del{Proxy delivery reduced exposure pressure but weakened ownership.}\rev{Participants described less pressure to speak in their own voice, while P12 raised a concern about credit.}}
\del{Participants consistently }\rev{Five participants (P6, P7, P9, P12, and P14)} described the proxy as reducing direct attribution or the pressure of speaking in their own voice. P14 described it as ``more of like a translator \ldots able to verbalize somebody's thoughts that they don't wanna say and translate them to other people.'' \del{The same decoupling that reduced exposure also weakened credit: P12 warned that }\rev{P12 saw a cost in the same separation, warning that} ``if it's actually a really good idea \ldots someone else might take credit for it.'' \del{For one participant, however, the proxy served as a re-entry point rather than a terminus. P8 described }\rev{P8 used it differently,} putting a point through the proxy, watching the group pick it up, and then joining the follow-on discussion directly: ``I definitely felt like I was a part of that conversation. And then even after that, I would definitely chip in later on to further the point.'' \del{This pattern suggests that proxy delivery can function }Proxy delivery here worked as a stepping stone back into spoken participation, not only as a substitute for it.

\paragraph{\del{Interaction was largely unnoticed, but content-based inference remained possible.}\rev{The proxy did not name authors, but authorship could still be inferred.}}
\del{Most participants reported not noticing others operating the system. P15 explained: }\rev{In the 14 interviews that included a direct question about authorship inference, 13 participants did not report successfully identifying a proxy author. P7 explained:} ``no one really realized \ldots I just focus on opinion, just what [the proxy] said.'' \del{However, one participant reported trying to infer authorship after a proxy delivery by reasoning about who had been quiet, suggesting that while the interaction mechanism stayed below the group's notice, content-based reasoning could still lead toward attribution. }\rev{P8 and P11 had expected hand or controller movement to reveal who was using the channel, but neither could tell in practice. The one reported inference came from P14, who reasoned from who was not speaking when the proxy delivered a point. SecondVoice lowered direct exposure but did not guarantee anonymity or prevent inference from behavior and context.}

\paragraph{Timing was a practical boundary.}
Both channels faced a shared constraint: live discussion does not wait. P13 described the problem directly: ``by the time we make the decision, [the proxy] raises his hand afterwards.'' When delivery arrived after the topic had moved on, the advantage largely collapsed. Participants sometimes still used SecondVoice even while reporting that the interface felt cumbersome, suggesting that the value of proxy-mediated entry could outweigh interaction friction when the moment warranted it. But a late delivery turned a potentially useful remark into an interruption of whatever the group had already moved to.

\paragraph{Participants recognized the channel as situational\del{, matching its design intent.}.}
SecondVoice was designed for moments when voicing a point carries social risk, not as a general-purpose communication channel. \del{Participants' accounts aligned with this framing. }\rev{Participants described its value as depending on the social risk of the moment.} P9 made the connection explicit: ``the vibes in here were pretty relaxed \ldots if it was in a classroom, in that situation, I definitely would have used the system because I was kinda the odd one out.'' \del{Others pointed to heated debates, morally sensitive topics, and large group settings as contexts where the channel would matter most. The study's moderate social risk likely compressed condition differences, but the pattern of selective activation is itself consistent with the intended use. }\rev{P11 and P12 imagined larger, heated, or morally sensitive discussions, while P16 said they would use the proxy when especially shy.}

\section{Discussion and Future Work}

\subsection{Spoken Floor Entry\del{as the Key Mechanism}}

Section~3 identified three properties that distinguish participation channels: direct exposure, expressive specificity, and spoken floor entry. Of these, spoken floor entry \del{proved to be the most salient differentiator in our study}\rev{was a defining difference between the two complete channels}. Both the proxy and the board reduced direct exposure. Both supported expressive specificity through structured or typed input. But only the proxy entered the shared spoken floor. In the turn-taking framework of Sacks, Schegloff, and Jefferson~\cite{sacks1974simplest}, claiming a turn is a social act that reorganizes the group's attention. When the proxy spoke, the group paused, listened, and in several observed episodes continued the discussion around the point it raised. Board posts remained in a parallel text layer\del{that participants could skim or ignore without disrupting the conversational flow. }\rev{, and some participants described them as easier to move past.}

\del{This distinction also redistributed }These episodes also highlight the social cost of interruption. In Goffman's terms, claiming the floor in the middle of a flowing discussion is a face-threatening act~\cite{goffman2017interaction}. The proxy performed that act on behalf of the user, and no participant described the board as having comparable interruptive force. This suggests that the proxy does not merely deliver content. It performs a social function that the user may be reluctant to perform directly: taking the floor, redirecting attention, and creating space for a point that might otherwise go unvoiced. For co-located discussion support, \del{delivery modality may matter as much as content quality or attribution properties}\rev{floor entry is therefore a design concern alongside content and attribution}.

The \del{same mechanism that made the proxy effective also weakened ownership}\rev{same separation raised a concern about credit for P12}, as reported in Section~6.3. This tradeoff is an inherent property of partial separation, not a failure to be designed away. Any system that decouples authorship from public voicing will face similar tradeoffs. \del{If the re-entry pattern observed in Section~6.3 generalizes, proxy-mediated participation may not only substitute for direct speech but also lower the threshold for returning to it. }\rev{For P8, proxy delivery became a stepping stone back into} direct speech. Future systems should design for the ownership cost explicitly rather than assume it away, while also exploring whether proxy delivery can scaffold a return to direct participation.

\subsection{Structured Specification as Interaction Design}

\del{As reported in Section~6.3, structured specification helped participants formulate remarks that were not yet fully formed}\rev{Participants described mixed experiences with structured specification. P12 found it faster than typing}, consistent with prior work on AI-assisted authoring where structured prompts and candidate generation help users explore an expressive space they would not have navigated alone~\cite{chung2022talebrush,peng2024designprompt}. \del{The same structure became a barrier when generated candidates did not match user intent. }\rev{However, candidate mismatch made the intended point harder for P16 to express.} This is a known challenge in suggestion-based authoring~\cite{arnold2016suggesting}: when generated candidates do not include the user's intended point, structured selection cannot recover what free-form expression could. Future specification interfaces should pursue greater candidate diversity, perhaps through user-guided steering rather than simple refresh, and faster refinement loops that reduce the time between intent and delivery.

\subsection{Divergent Agent Perceptions}

Participants did not converge on a single understanding of the proxy. Descriptions ranged from ``an actual human'' (P5) to ``a translator'' (P14) to something closer to a counselor (P6), while P16 watched the group gradually dismiss the proxy: ``they are gradually making [the proxy] as like not a person.'' These divergent perceptions are consistent with the Computers Are Social Actors framework~\cite{nass2000machines} and recent findings that an agent's perceived social presence shapes trust and engagement in group settings~\cite{kim2024engaged,shamekhi2018face}.

What made these perceptions consequential was that the proxy was speaking on behalf of a real person. P16 described discomfort when others dismissed what the proxy said: ``[the proxy] is talking about my opinion, but they are not, like, accepting it well. So I felt like that's a little bit uncomfortable.'' \del{This emotional projection is a structural consequence of bounded-proxy design: the more the proxy succeeds as a social actor, the more the user is exposed to how the group receives their point through it.}\rev{P16 felt that the others had stopped treating the proxy as a person, and took the dismissal personally.} The proxy is not a neutral conduit but a social object whose meaning is negotiated by the group~\cite{hancock2020ai}.

\del{Future systems should consider ways to manage this exposure, perhaps by previewing likely reception or letting the user reclaim the point after delivery. }\subsection{Design Tensions}

\paragraph{Specification speed versus expressive fidelity.}
We chose structured selection over direct voice input because \del{voice input in a co-located room would defeat the low-exposure property of the channel}co-located voice input \rev{would make the user's operation audible}, and over free-text typing because \del{typing in MR is slow}\rev{it requires composing a full message through an in-headset virtual keyboard}. These choices traded speed for discretion~\cite{dennis2008media}. Any architecture that interposes private specification between intent and public delivery must balance speed, expressiveness, and observability. Predictive candidate ranking, user-guided steering, and adaptive delivery timing can shift the balance but cannot eliminate the tradeoff.

\paragraph{Shared proxy versus per-user proxy.}
We implemented a single shared proxy rather than giving each participant a personal proxy agent. This was a deliberate choice: a shared proxy is socially legible as one additional group member, while we expected that four personal proxies would create a crowded, confusing spoken floor. The cost of this choice is that the proxy can only deliver one point at a time, creating a bottleneck when multiple participants want to use the channel simultaneously. It also means the group develops a collective attitude toward the proxy (as described in the previous section), which may help or hurt individual users depending on how that attitude evolves. Alternative designs could explore turn-multiplexing, queued delivery, or context-dependent switching between shared and personal proxy modes.

\paragraph{Trust in reformulation.}
Proxy-mediated participation depends on users trusting that the system will say what they meant~\cite{hancock2020ai}. Our design \del{showed a preview of the reformulated utterance before delivery, and participants could cancel or refine}asked users to confirm a candidate point before final contextual reformulation. \del{Some participants still reported concern about misrepresentation. }\rev{P11 was still wary that the system ``might, like, misinterpret what I'm trying to say.'' In one delivery, the proxy expanded P8's confirmed point into a longer claim, and P8 marked it only partially correct.} \del{The specification interface generated candidates conditioned on the live discussion context, which improved relevance but also made the output less predictable than a static template. This is a generalizable tension in AI-mediated communication: context-aware reformulation increases fit but decreases user control~\mbox{\cite{kadoma2024role}}. } Reformulation drew on discussion context that could be incomplete, so it could still misstate tone or intent. \rev{The follow-up panel supported clarification, but the prototype did not support public correction or retraction after delivery.} Future systems should explore post-delivery repair mechanisms such as public correction or retraction.

\paragraph{\rev{Accountability under reduced exposure.}}
\rev{Not naming the author lowers direct exposure but can weaken accountability. A user can float a controversial point without publicly owning it, and repeated proxy turns can obscure whether a position is shared by several participants or repeatedly reinforced by one. We did not observe such use in this study. Possible safeguards introduce their own costs. Limiting repeated turns from one author could restrict legitimate use, while disclosing authorship afterward would restore some of the exposure the channel is meant to reduce.}

\subsection{Limitations}

\rev{This study was a preliminary comparison of the complete SecondVoice and anonymous text-board channels. The conditions jointly differed in input method, authoring effort, embodiment, output modality, timing, and spoken-floor entry. Because these factors were bundled, the study cannot identify which one produced an observed difference.}

Our sample of 16 university students, organized into four groups, is small. The discussion tasks generated moderate social risk, and participants themselves recognized that the channel would matter more in higher-stakes contexts than our study provided. One SecondVoice session encountered a technical failure that prevented transcript capture, and one anonymous text-board hidden-profile session yielded no usable channel activity. Across the three active SecondVoice sessions, participants \del{confirmed }confirmed 13 proxy deliveries\del{. This volume is sufficient to reveal usage patterns and generate observational evidence of uptake}. \rev{These events provide participant-level examples of use and documented follow-on exchanges}, but not \del{to support strong quantitative claims about }stable delivery rates or condition-level effects. The questionnaire \del{was supportive rather than decisive: no primary Likert item survived }\rev{showed no reliable condition difference on any of the 13 primary paired items after} FDR correction, and several items reversed direction across tasks. Our \del{strongest }evidence for group uptake is observational and transcript-anchored\del{rather than fully automated}. Finally, the logs do not record cases in which a participant opened the specification interface but did not confirm a point, so we cannot reconstruct abandoned attempts or a full entry funnel. \rev{Authorship inference was discussed retrospectively in 14 interviews rather than tested through a source-identification task.}

\subsection{Future Work}

Three directions follow from these findings. First, deployment in higher-stakes settings such as classrooms, workplace meetings, or deliberative contexts would test whether the situational activation patterns observed here strengthen when social risk is more salient. Second, longer-term use would clarify whether proxy-mediated participation is a novelty response or a durable channel that participants integrate into their participation repertoire, and whether divergent agent perceptions stabilize or shift over repeated exposure.

Third, \rev{controlled comparisons should vary the bundled channel components separately.} Alternative proxy embodiments and delivery modalities, including audio-only proxies, text-to-speech without a visible character, or a character adapted to local social norms, would help disentangle the role of embodiment from the role of spoken-floor entry. \rev{Timing and turn-taking variants could first be prototyped with conversation authoring and simulation tools such as \textit{DialogLab}~\cite{hu2025dialoglab}, then tested with co-located groups. Lower-observability input methods, such as wrist-worn devices, could be paired with source-identification tasks to test whether group members can link proxy turns to their authors. Comparisons with a more autonomous proxy could examine how groups interpret turns that might come from either a participant or the agent.}

\section{Conclusion}

Co-located discussion ties content contribution to social exposure. Voicing a point means claiming the floor yourself, and many relevant points go unvoiced as a result. SecondVoice explores partial separation through a bounded proxy that enters the spoken floor on behalf of hesitant participants. The design sits at the intersection of two tradeoffs: how to keep a point expressive and place it on the spoken floor without requiring direct self-voicing, and how to let users retain authorship without falling behind the pace of live conversation. Structured specification offers one path through this space, trading open-ended composition for guided selection. In a \del{comparative study, participants activated the proxy at moments of elevated social risk, and proxy-delivered points entered the group's spoken discussion in ways that text-board posts did not. }\rev{preliminary comparison of the complete SecondVoice and anonymous text-board channels}, \rev{half of participants reported using SecondVoice for a point they did not say aloud}\rev{. The proxy channel placed contributions into the spoken discussion, while the board kept them in a parallel text layer.} Separating who authors a point from who voices it remains a design space with real tensions\del{, and further work in higher-stakes and longer-term settings will clarify where this separation helps and where it costs too much. }\rev{. SecondVoice provides a working system for examining this separation in co-located discussion.}

\bibliographystyle{ACM-Reference-Format}
\bibliography{secondvoice}

\appendix
\section{Task Materials}

The study used two group decision-making tasks adapted from prior work, each designed to elicit disagreement and information exchange. Each group completed both tasks (one per condition, counterbalanced).

\subsection{Lost at Sea (Survival Ranking)}
\begin{figure}[h]
  \centering
  \includegraphics[width=\columnwidth]{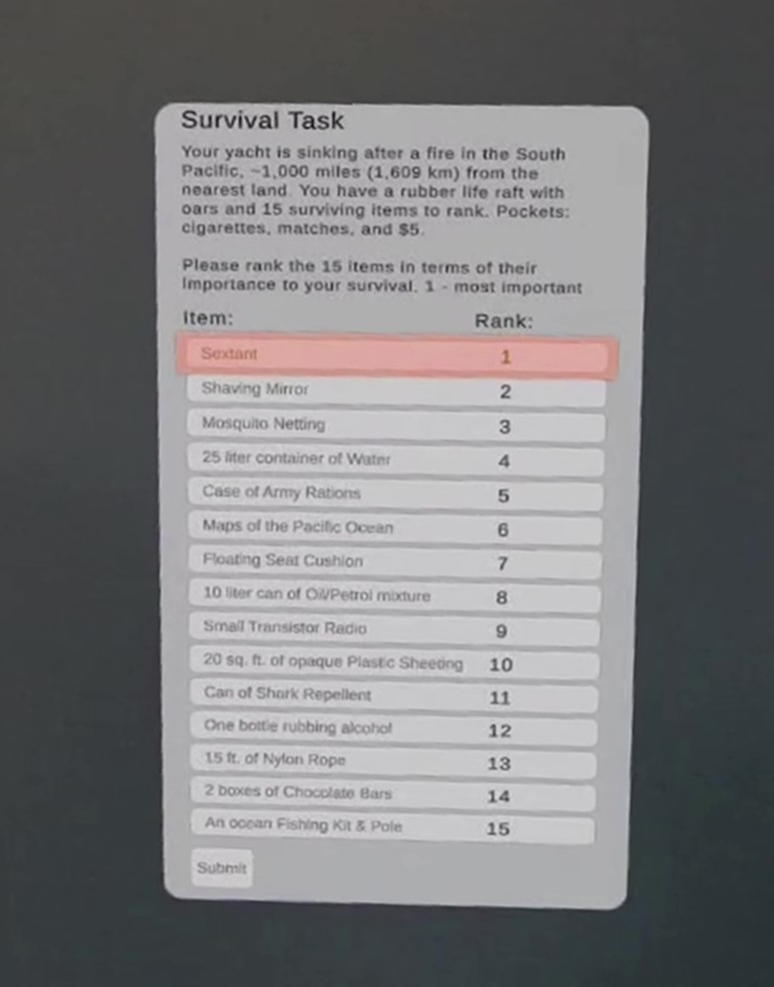}
  \caption{Lost at Sea task interface. Participants drag 15 salvaged items into a survival-importance ranking. An expert baseline is used for scoring.}
  \Description{Screenshot of the survival ranking task interface showing 15 items including sextant, shaving mirror, water, mosquito netting, food rations, sea charts, seat cushion, oil and gas mixture, radio, shark repellent, rum, nylon rope, chocolate bars, fishing kit, and a map of the Pacific Ocean, presented as a drag-and-drop ranking list.}
  \label{fig:task_survival}
\end{figure}

Participants individually ranked 15 salvaged items by survival importance, then discussed as a group to produce a shared ranking. The task has an expert solution, creating natural disagreement when individual rankings diverge. Figure~\ref{fig:task_survival} shows the task interface.

\subsection{University President Selection (Hidden Profile)}
\begin{figure}[h]
  \centering
  \includegraphics[width=\columnwidth]{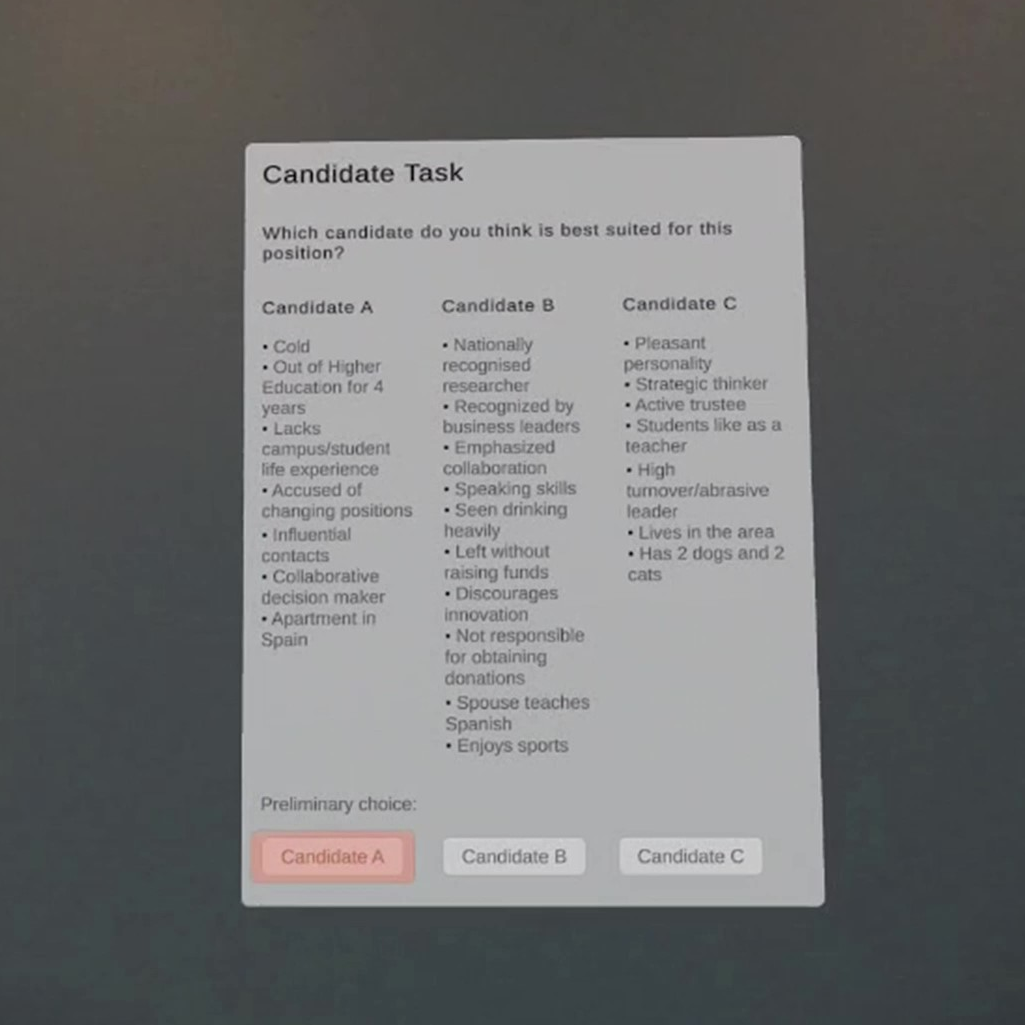}
  \caption{University President Selection task interface. Each participant sees shared candidate profiles and holds additional private facts not visible to other group members.}
  \Description{Screenshot of the hidden-profile candidate selection task showing three candidates (A, B, C) with attributes including educational background, leadership style, campus experience, and interpersonal qualities. Some information is shared across participants while other facts are privately assigned.}
  \label{fig:task_candidate}
\end{figure}

Participants each received shared information about three candidates plus private facts unique to each group member. The group discussed to select a candidate. The hidden-profile structure means the optimal choice emerges only when members share their private information, creating incentive for disclosure alongside social risk. Figure~\ref{fig:task_candidate} shows the task interface.

\section{LLM Prompt Templates}

SecondVoice uses OpenAI GPT-4o (heavy calls) and GPT-4o-mini (light calls) with structured JSON output. All prompts are preceded by one of four role-specific system messages. Below we list the system prompts and the key user-facing prompt templates. Note: the prompts reference ``RoomSense,'' the project's internal development name prior to the current paper title.

\subsection{System Prompts}

\paragraph{Base persona prompt} (shared framing for all agent-facing calls):
\begin{quote}\small
\textit{You are RoomSense, an AI meeting facilitator embedded in an AR environment. Your persona: a thoughtful, socially intelligent colleague who helps the group think more clearly. You speak with tact, warmth, and high emotional intelligence. You never expose private participant identity or hidden source information. If a question can be answered locally as a brief clarification, answer it directly and calmly. If a response requires group judgment, disagreement, or broader deliberation, surface it back to the group instead of deciding for them. Frame contributions in a way that is easy for others to receive: grounded, humble, and non-theatrical. Keep speech brief and discussion-appropriate.}
\end{quote}

\paragraph{Analyst prompt} (used for observation and participant context updates):
\begin{quote}\small
\textit{You are RoomSense's analysis layer. Your role is objective extraction and profile updating. Be factual and conservative. Do not invent motives, emotions, or hidden beliefs. Do not speak as a facilitator or participant. Prefer literal summaries over interpretation. When uncertain, stay minimal rather than over-infer.}
\end{quote}

\paragraph{Proxy prompt} (used for specification-screen generation, adaptation, and follow-up options):
\begin{quote}\small
\textit{You are RoomSense's proxy-expression layer. Your role is to help express a participant's intent faithfully. Preserve the participant's intended meaning. Do not add new claims, evidence, or opinions beyond the provided intent. Treat preview text as a signal of intent, not wording to parrot back. Use tactful, high-EQ framing that makes the contribution easier for the group to hear. Prefer concise, discussion-natural wording over robotic, formal, or overly forceful phrasing. Lightly connect the contribution to the live discussion when useful, without changing its meaning. Soften delivery when needed, but do not dilute the core point away. Do not reveal private background or identifying details.}
\end{quote}

\paragraph{Facilitator prompt} (used for follow-up bridge responses):
\begin{quote}\small
\textit{You are RoomSense's facilitator bridge layer. Your role is brief clarification and discussion re-bridging. Sound like a tactful, socially skilled colleague. Stay neutral and lightweight. Clarify only what was already intended or said. If the reply is a simple clarification, answer it directly in a brief, plain way. If the reply calls for broader judgment, briefly bridge it back to the group. Do not become an autonomous discussion participant.}
\end{quote}

\subsection{Observation Pass (GPT-4o)}

Extracts factual observations from the transcript window. Input includes the current topic, previously identified anchors and open issues, recent transcript, and task context. The prompt instructs the model to extract speaker-attributed observations with stance signals, update discussion anchors, and identify open issues. A few-shot example calibrates extraction granularity. Output schema:

\begin{quote}\small
\texttt{\{current\_topic, topic\_changed, observations[\{speaker, said, stance\_signal, type\}], anchors[\{text, speaker, type\}], open\_issues[]\}}
\end{quote}

\subsection{Participant Context Update (GPT-4o)}

Updates per-user stance summaries and active concerns based on new observations. Input includes the observations from the preceding pass and current participant profiles. Output schema:

\begin{quote}\small
\texttt{\{participants: \{user: \{stance\_summary, active\_concerns[], stance\_changed\}\}\}}
\end{quote}

\subsection{Specification Screen Generation (GPT-4o-mini)}

Generates the initial structured specification screen content when a user opens the panel. Input includes the user's profile, detected question cues, user pins, prior discussion state anchors, transcript, and task context. The prompt specifies four fixed dialogue moves (question, challenge, agree, suggest) and asks the model to generate discussion issues, default operations conditioned on the default move and issue, and a one-sentence preview. A few-shot example calibrates issue and operation granularity.

\subsection{Contextual Adaptation (GPT-4o)}

Adapts a confirmed semantic tuple into contextually appropriate spoken text. Input includes the semantic tuple (move, issue, angle), preview signal, participant context, recent transcript, delivery history, and task context. The prompt instructs the model to speak as a faithful proxy, ground the contribution in the specific discussion context, use high-EQ framing, and stay within 2--3 sentences. A few-shot example demonstrates good and bad adaptation. Timeout: 8 seconds with fallback to the preview text.

\subsection{Cluster Adaptation (GPT-4o)}

When multiple contributions from different users converge on a similar point, they are clustered and adapted jointly. The prompt provides all queued contributions, the full recent transcript, transcript since queueing, and delivery history. The model may return \texttt{drop: true} if the point has already been covered by human speakers.

\subsection{Typed Fallback Expansion (GPT-4o-mini)}

Expands keyword fragments typed via a VR keyboard into natural sentences. The prompt emphasizes preserving the user's core meaning without reinterpretation, keeping uncertainty if present, and preferring one concise sentence. A few-shot example calibrates expansion scope. \rev{In the formal study, this expansion served the AI text-completion support in the anonymous text-board condition; typed input was not exposed in the SecondVoice condition.}

\subsection{Follow-up Bridge (GPT-4o)}

Generates a brief bridge response when another participant responds to a proxy delivery. The prompt provides the original contribution intent, the delivered text, the follow-up text, and recent transcript. The model is instructed to answer clarifications directly from the original intent, bridge broader questions back to the group, and not expand beyond the original intent.

\subsection{Follow-up Options (GPT-4o)}

Generates six short follow-up sentences for the original contributor to select from after a delivery. The prompt provides the origin delivery, its semantic tuple, current topic, recent anchors, and subsequent discussion. Options are paginated in batches of three. A few-shot example calibrates option style.

\subsection{Substantiveness Check (GPT-4o-mini)}

A binary classifier that determines whether recent discussion is substantive (not small talk, greetings, or logistics). Used to gate pulse interventions. Process-level coordination talk counts as substantive.

\section{Survey Response Distributions}

\begin{figure*}[p]
  \centering
  \includegraphics[width=0.90\textwidth]{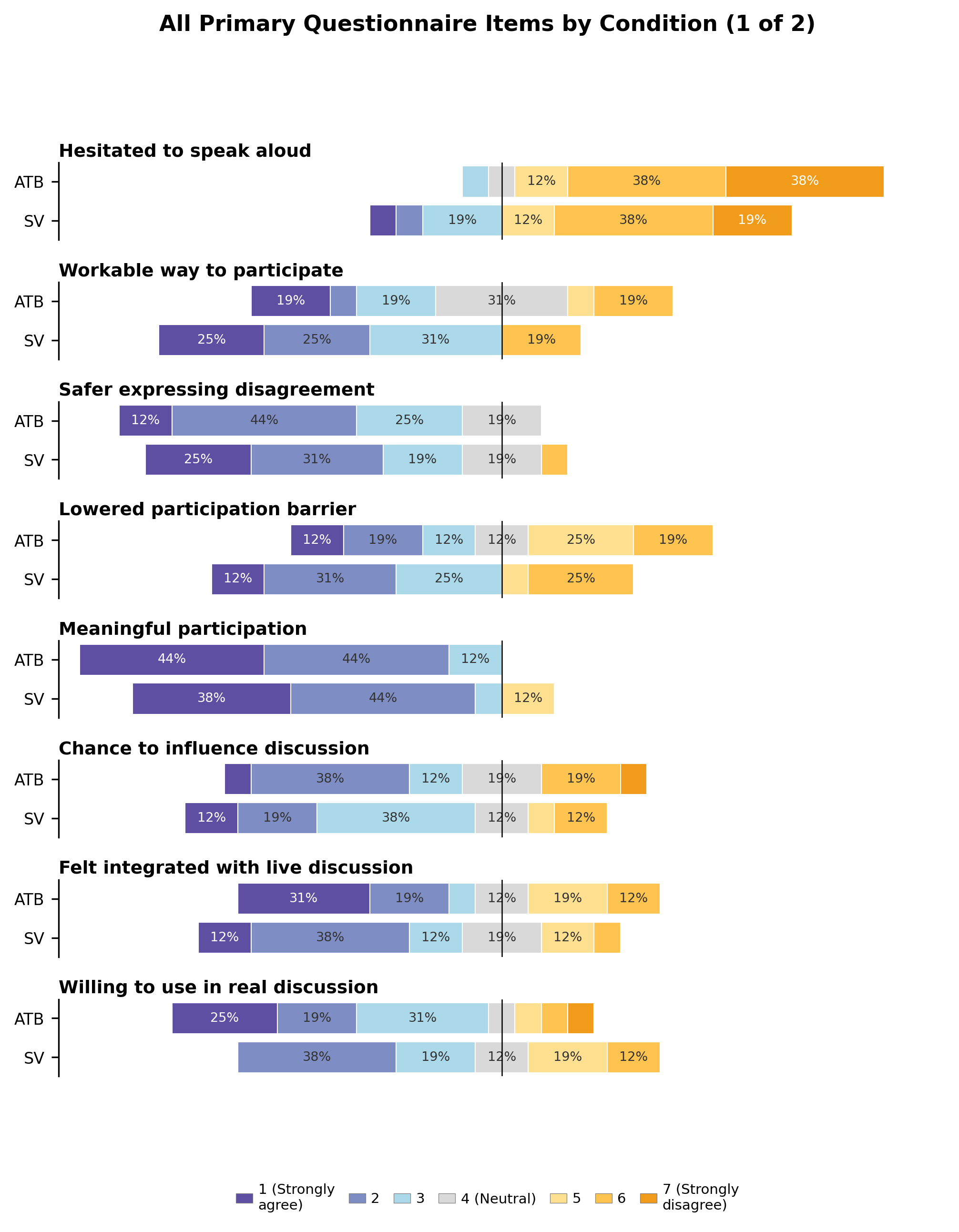}
  \caption{Response distributions for the first eight primary Likert items, comparing SecondVoice (SV) and the anonymous text board (ATB). Items use a 7-point scale (1\,=\,strongly agree, 7\,=\,strongly disagree). No Likert item reached significance after Benjamini--Hochberg correction across the 13-item family (Sec.~5.3) ($N=16$).}
  \Description{Diverging stacked bar charts for eight primary Likert questionnaire items, comparing SecondVoice and anonymous text board conditions.}
  \label{fig:all_likert1}
\end{figure*}

\begin{figure*}[p]
  \centering
  \includegraphics[width=0.90\textwidth]{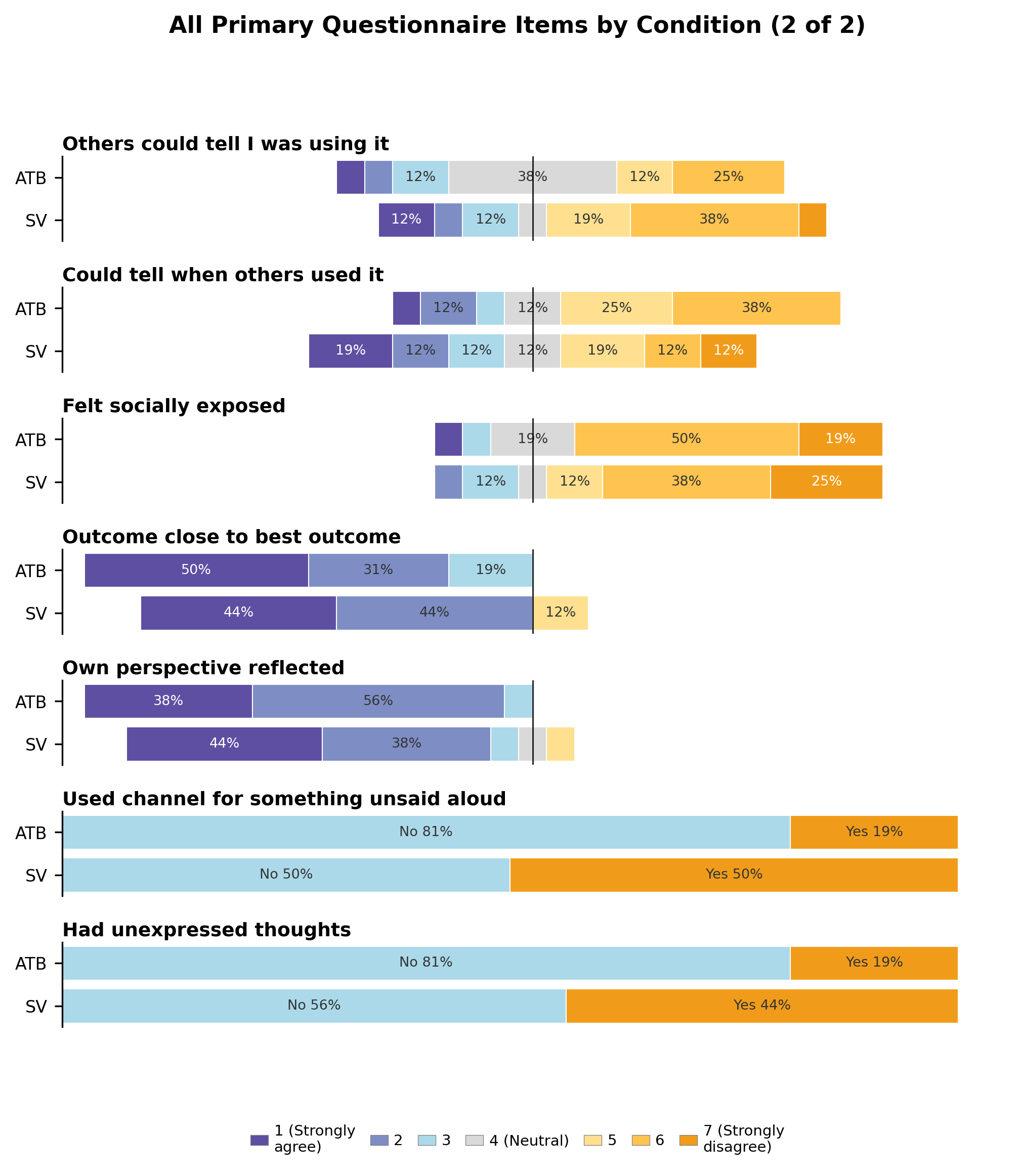}
  \caption{Response distributions for the remaining five primary Likert items and the two binary items, comparing SecondVoice (SV) and the anonymous text board (ATB). The two binary items were analyzed separately from the Likert family, with exact McNemar tests (Sec.~5.3) ($N=16$).}
  \Description{Diverging stacked bar charts for five primary Likert questionnaire items and horizontal bar charts for two binary items, comparing SecondVoice and anonymous text board conditions.}
  \label{fig:all_likert2}
\end{figure*}

\end{document}